\documentclass[letterpaper, 10 pt, conference]{ieeeconf}  

\IEEEoverridecommandlockouts                              

\newcommand{\mbb}[1]{\mathbb{#1}}

\usepackage{graphics} 
\usepackage{epsfig} 
\usepackage{times} 
\usepackage{flafter}   
\usepackage{placeins}  
\usepackage{amsmath} 
\usepackage{amssymb}  
\usepackage{bm} 
\usepackage{xcolor}

\newcommand{\Rn}{\mathbb{R}}
\usepackage{subcaption}
\title{\LARGE \bf
Risk-Aware Kinodynamic Motion Planning Under Uncertainty For Safe Navigation on Planetary Environments
}
\author{Sachin Sunil Kelkar$^{1}$, Tanmay Dokania$^{1}$, Yashwanth Kumar Nakka$^{1}$ 
\thanks{$^{1}$Authors are with the Daniel Guggenheim School of Aerospace Engineering at the Georgia Institute of Technology, Atlanta, GA, 30332 USA. Email: [skelkar37, tdokania3, ynakka3]@gatech.edu
}}

\begin{document}
\maketitle
\thispagestyle{empty}
\pagestyle{empty}

\begin{abstract}
For autonomous space exploration, robotic agents need to perform motion planning in which environmental interactions may be unknown. Learning these interactions, such as terrain mechanics for wheeled robots, can introduce uncertainties that lead to risky motion plans and potentially hazardous operations or mission failures. Moreover, uncertainties induced by perception-based systems can exacerbate the problem of safe motion planning. In this letter, we address the problem of performing cost-optimal kinodynamic motion planning with risk awareness. We approach this in two steps. First, a sampling-based planner (AO-RRT) generates a dynamically feasible and risk-aware asymptotically cost-optimal trajectory. Second, we formulate motion planning as a nonlinear optimization problem and solve it using sequential convex programming (SCP), using the AO-RRT trajectory as an initial solution. By quantifying risk using conditional value-at-risk (CVaR), we demonstrate a reduction in risk by over $\sim$97\% across trajectories in simulation and hardware experiments.
\end{abstract}

\section{Introduction}

Risk-aware motion planning is the problem of constructing a feasible trajectory for a robot to follow around obstacles while accounting for uncertainties arising from various sources. It is critically important for autonomous space exploration, where robotic agents traverse uncertain environments, with actuation degradation and unmodeled terrain dynamics. A planner that does not account for these uncertainties can lead to mission-critical failure or hazardous operations \cite{majumdar2020risk}. In previous works, uncertainty in the environment and sensor model are considered to generate risk-aware geometric trajectories \cite{dixit2024step} or modeled as a function of terrain features \cite{10606099}. A two-step approach in \cite{8767973} tracks a risk-neutral geometric plan with a CVaR-constrained controller. We consider quantifying the uncertainty in the learned dynamics model used by the planner to generate a risk-aware initial trajectory. 

To address these problems, our contributions are: (1) Quantify uncertainty in the learning-based dynamics model using conformal prediction, (2) Formulate AO-RRT with closed-loop risk computation as cost, (3) Cast the nonlinear problem as locally convex and optimize for CVaR and control cost in the given AO-RRT homotopy, guaranteeing the nonlinear dynamics are followed. 

\begin{figure}[t]
  \centering
  \begin{subfigure}{0.48\columnwidth}
    \centering
    \includegraphics[width=\linewidth]{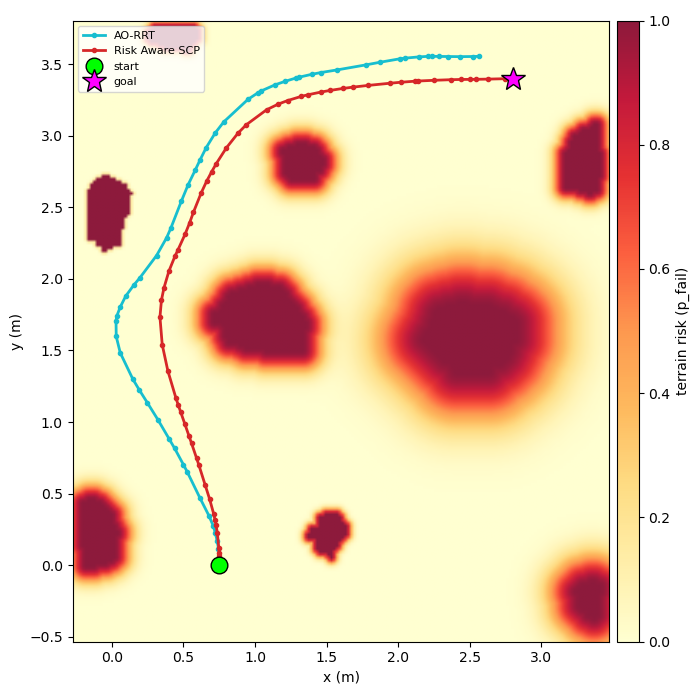}
    \caption{AO-RRT with risk cost}
    \label{fig:pair-a}
  \end{subfigure}
  \hfill
  \begin{subfigure}{0.48\columnwidth}
    \centering
    \includegraphics[width=\linewidth]{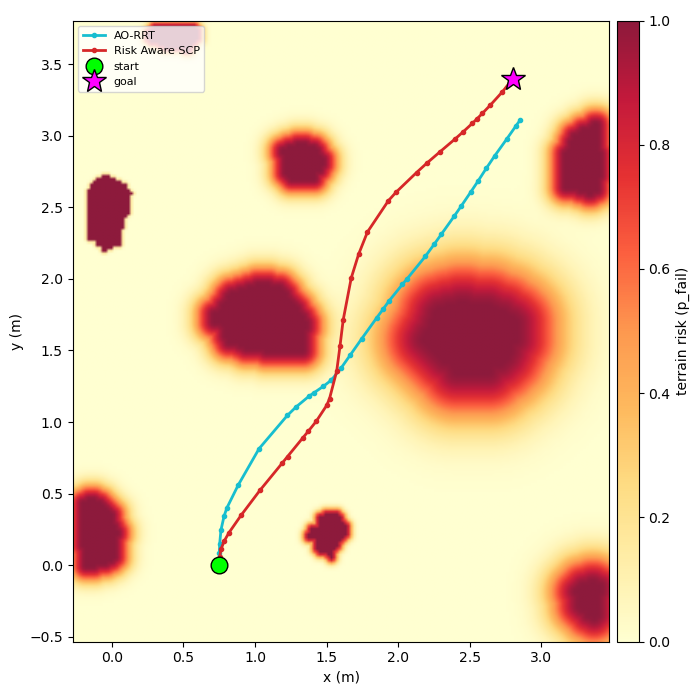}
    \caption{AO-RRT without risk cost}
    \label{fig:pair-b}
  \end{subfigure}
  \caption{For the same initial conditions and obstacle uncertainty space, AO-RRT with risk cost generates an initial trajectory solution for Sequential Convex Programming that has less joint collision risk}
  \label{fig:sim-comp}
\end{figure}

\section{Dynamics}
We consider the following uncertain dynamics model, \cite{lupu_magicvfm_2025}. 
\begin{equation}\label{eq:syseq}
    \dot{x} = f(x,u,t) +d(x,u),
\end{equation}
here $x\in \mbb{R}^{n}, u\in \mbb{R}^{m}, d(x,u) \in \mbb{R}^{n}$ are the system state, control input, and the unmodeled dynamics residue term. We model the system as a first-order velocity-controlled model with states,  $v =[v_x,\omega]^\top$, and control $u = [v_\text{cmd},\omega_\text{cmd}]^\top$. We learn and approximate the residual term using a Lipschitz-continuous neural network as $d(v,u)  \approx \Phi (\mathbf{v}){\mathbf{u}}$. 

To quantify the uncertainty of the neural network's output and incorporate it into motion planning, we use conformal prediction \cite{lindemann_safe_2023} by computing the error between the predicted residual dynamics and the ground truth measured with a motion capture system. Hence, we obtain the set $\mathcal{M}$,

\begin{equation}
    \mathcal{M} = \{l_\text{min},\dots ,l_\text{max}\} \times \{e_\text{min},\dots,e_\text{max}\} \in \mathbb{R}^{l\times e},
    \label{eq:dis set}
\end{equation}
here $l,e_\text{min/max}$ are the minimum and maximum error in the linear and angular acceleration residual terms, computed at the $95\%$ confidence level. 

\begin{figure*}[t]
    \centering
    \begin{subfigure}[b]{0.49\textwidth}
        \includegraphics[width=\linewidth]{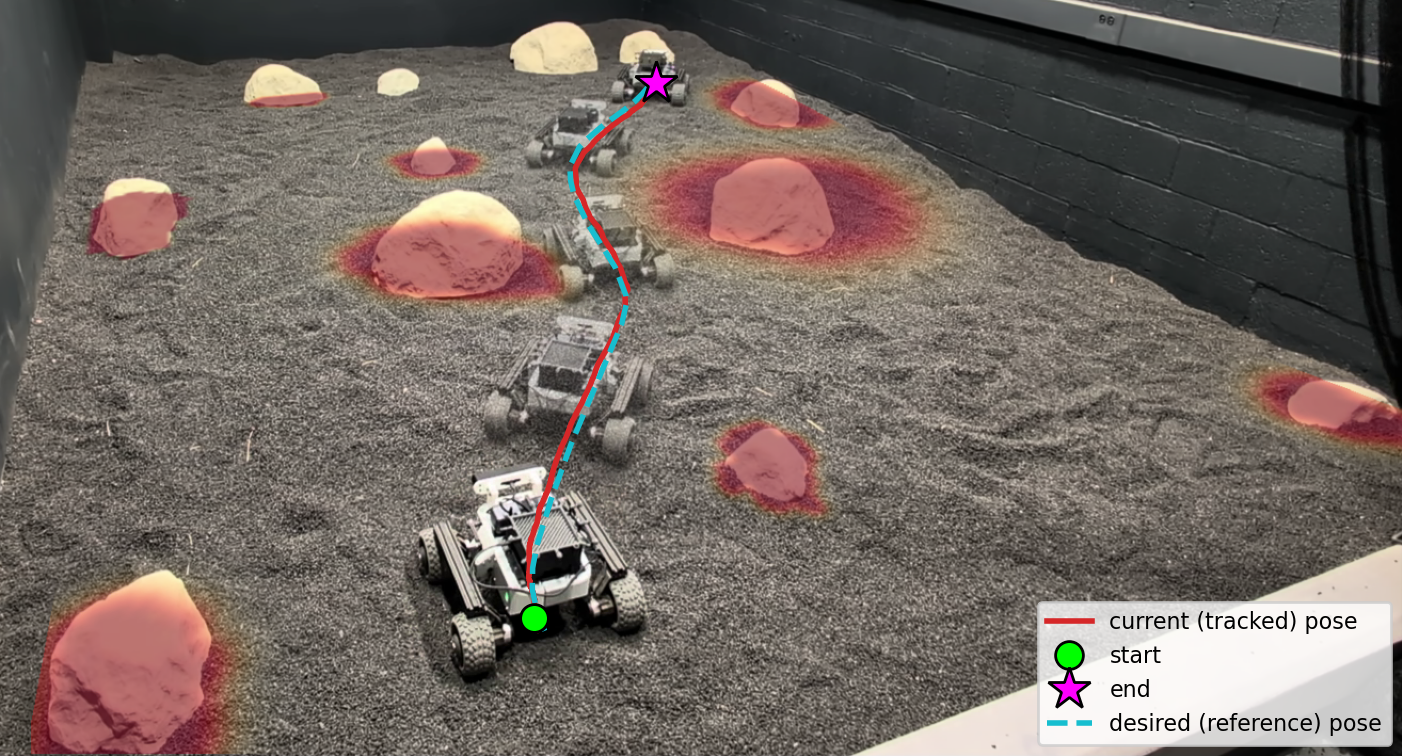}
        \caption{Without Risk-aware Cost}
        \label{fig:aorrt_norisk}
    \end{subfigure}
    \hfill
    \begin{subfigure}[b]{0.49\textwidth}
        \includegraphics[width=\linewidth]{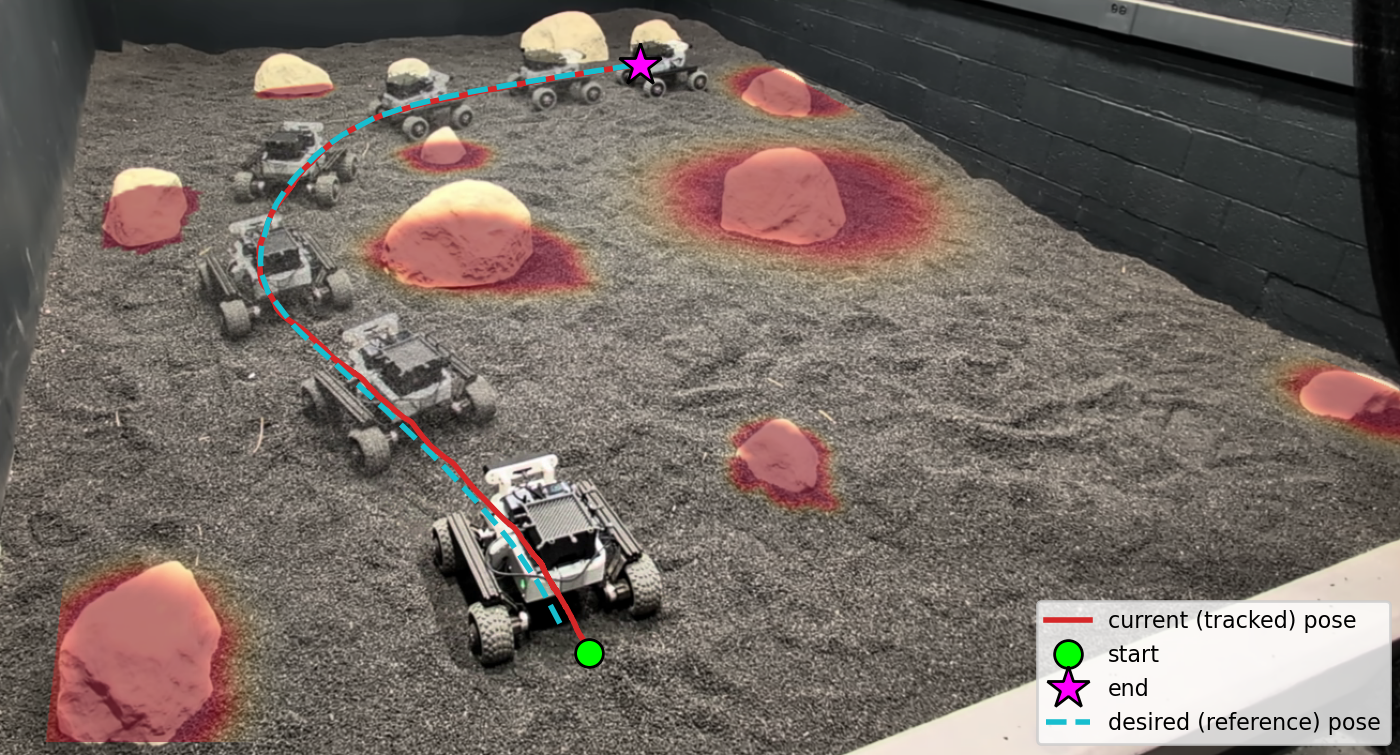}
        \caption{With Risk-aware Cost}
        \label{fig:aorrt_risk}
    \end{subfigure}
    \caption{Experimental demonstration of the trajectory tracking for the generated trajectories on the Leo rover. The red region around the obstacles represents the uncertainty around them. When jointly optimizing over the cost and risk, we demonstrate that the resulting trajectory has a lower joint collision risk.}
    \label{fig:aorrt_comparison}
\end{figure*}

\subsection{Uncertain Obstacle Space}

We quantify proximity to the boundaries of uncertain obstacles as a risk field over $\mathbb{R}^2$. For obstacle $i$ with a user-specified boundary-uncertainty scale $\sigma_i$, and signed distance $d_i(x)$ from its nominal boundary, the risk is $r_i(x) = \exp\!\big(-d_i(x)^2 / 2\sigma_i^2\big)$. Treating each obstacle as an independent failure mode, the per-obstacle scores are fused into a single risk field, or $\mathrm{RiskMap}$,
\begin{equation}
\mathrm{RiskMap}(x) = 1 - \prod_{i=1}^{n}\big(1 - r_i(x)\big),
\label{eq:riskmap}
\end{equation}
which is bounded in $[0,1]$, monotone in each $r_i$, and saturates as risk accumulates across nearby obstacles. This field serves as the risk map against which planned trajectories are evaluated.

Additionally, as a soft risk penalty for risk-aware AO-RRT, we inflate each obstacle's signed distance field $\mathrm{sdf}_i(x)$ outward by a CVaR margin,$ F_i(x) = \mathrm{sdf}_i(x) - \kappa(\alpha)\,\sigma_i,
\kappa(\alpha) = \frac{\phi\big(\Phi^{-1}(1-\alpha)\big)}{\alpha},
$ where $\kappa(\alpha)$ is the CVaR multiplier of a standard normal at tail
probability $\alpha$ ($\phi,\Phi$ the standard normal PDF and CDF). The
risk-inflated SDF is the tightest margin across all $n$ obstacles,
\begin{equation}
\mathrm{RiskSDF}(x) = \min_{i=1,\dots,n} F_i(x),
\label{eq:risksdf}
\end{equation}

\subsection{AO-RRT Implementation}
We generate a dynamically feasible plan using the Asymptotically Optimal Rapidly Exploring Random Trees (AO-RRT) algorithm \cite{hauser_asymptotically_2016}.

The nodes of the tree are states of the rover, $\mathbf{x} = [x,y,\theta,v_x,\omega]^\top$, which are extended by applying a randomly sampled controlled input, $\mathbf{u} =[v_\text{cmd},\omega_\text{cmd}]$, to each node.

To account for uncertainty in residual dynamics prediction, each edge's open-loop reference is tracked in closed loop over a horizon of $n_t$ steps. At every step, the controller compares the rover's actual state to the reference, computes the tracking control, and advances the state under the disturbed dynamics from \eqref{eq:dis set}. States in this trajectory are $x^{(i)}_{k,m}$, the $i^{\text{th}}$ state in the tracked trajectory of the $k^{\text{th}}$ AO-RRT node under the $m^{\text{th}}$ disturbance from \eqref{eq:dis set}, with the rollout vectorized in $\texttt{JAX}$ across all $M$ disturbances in the grid. We empirically estimate the CVaR of soft collision-constraint violation across this ensemble as the average violation over the worst $\alpha$-fraction of scenarios in $\mathrm{RiskSDF}$ \eqref{eq:risksdf}. Using this penalty as the AO-RRT edge cost, we account for dynamics uncertainty and obstacle position uncertainty in a single shot. Due to the sequential nature of cost propagation in the AO-RRT tree, the risk measure is time-consistent \cite{singh_framework_2019}.

\subsection{SCP Implementation}
The trajectory produced by the AO-RRT implementation is guaranteed to be dynamically feasible, but not guaranteed to be smooth in controls and states. Using sequential convex programming, we linearize the problem at each iteration and solve the convex subproblem in the trust region. By casting risk and control as objectives, we jointly minimize them at each state of the trajectory.

The pointwise terrain risk along the tracked sub-steps is aggregated as
\begin{equation}
{R}_{k,m} = \frac{1}{\beta}\log\!\left(\sum_{i} \exp\!\big(\beta\,\text{RiskMap}({x}_{k,m}^{(i)})\big)\right)
\label{eq:risk}
\end{equation}

We cast risk optimization as an objective using the Rockafellar-Uryasev CVaR epigraph \cite{rockafellar2000optimization}. For our problem, we define 
\begin{equation}
    \text{CVaR}_\alpha(R_k) = \min_{\tau_k}\ \Big\{ \tau_k + \frac{1}{\alpha M}\sum_{m=1}^{M} \eta_{k,m} \Big\}.
    \label{eq:cvar}
\end{equation}
The bar notation defines the initial solution of the SCP iteration, whereas variables without bar are the SCP solution. The slack variable $\eta\geq 0$ is defined as $\eta_{k,m} \ge \tilde{R}_{k,m} + g^s_{k,m}\, x_k + g^u_{k,m}\, u_k - \tau_k,$, where $\tilde{R}_{k,m} = R_{k,m} - g^s_{k,m}\bar{x}_k - g^u_{k,m}\bar{u}_k $ . Here, $g^s,g^u$ are gradients of risk map with respect to the state and control, computed exactly by autodiff in $\texttt{JAX}$. The matrices are defined as $g^s_{k,m} = \frac{d R_{k,m}}{d s_k} \in \mathbb{R}^{1\times5}, 
g^u_{k,m} = \frac{d R_{k,m}}{d u_k} \in \mathbb{R}^{1\times2}.$

We define the convex problem with decision variables, $x\in\mathbb{R}^{(K+1)\times5}$, $u\in\Rn^{K\times2}$, $\tau\in\Rn^{K}$, $\eta\in\Rn^{K\times M}_{\ge0}$, split virtual controls $\nu^{+},\nu^{-}\in\Rn^{K\times5}_{\ge0}$ (defect slack), terminal slacks $\sigma_T^{+},\sigma_T^{-}\in\Rn^{5}_{\ge0}$, and SDF buffers $\sigma\in\Rn^{K+1}_{\ge0}$.

The convex objective is defined as:
\begin{equation}
\begin{aligned}
\min\quad & \underbrace{w_u \sum_{k} \|u_k\|_R^2}_{\text{effort}} 
+ \underbrace{w_{\text{risk}} \sum_{k}\!\Big(\tau_k + \tfrac{1}{\alpha M}\!\sum_{m}\eta_{k,m}\Big)}_{\text{CVaR (epigraph, Eq.~\eqref{eq:cvar})}} \\
& + \underbrace{w_{\text{term}}\mathbf{1}^\top(\sigma_T^{+}\!+\!\sigma_T^{-})}_{\text{terminal }\ell_1} 
+ \underbrace{w_\nu \mathbf{1}^\top(\nu^{+}\!+\!\nu^{-})}_{\text{defect }\ell_1} \\
& + \underbrace{w_{\text{sdf}}\mathbf{1}^\top\sigma}_{\text{SDF buffer}} 
+ \underbrace{w_{\text{arc}}\!\sum_{k}\|p_{k+1}\!-\!p_k\|_2^2}_{\text{arc length}}
\end{aligned}
\label{eq:objective}
\end{equation}

\subsubsection*{Constraints}
The value of SDF $h$ at point in space, $\bar h_k \triangleq h_{\text{sdf}}(\bar p_k)$ and $\nabla_k \triangleq \nabla_p h_{\text{sdf}}(\bar p_k) \in \mathbb{R}^{1\times2}$ for the SDF and its spatial gradient at the state in the current iteration, and $p_k \triangleq [x_k^{(1)},x_k^{(2)}]^\top$ for the position block of $x_k$. Gradients here are computed for true SDF, with deterministic obstacle boundaries. All constraints below are affine in the decision variables and hold for every state $k$ and disturbance scenario $m$, with all slacks nonnegative.
\begin{subequations}
\label{eq:cons}
\begin{align}
  x_0 &= x_{\text{start}}, \quad
    x_K = x_{\text{goal}} + \sigma_T^{+} - \sigma_T^{-} \label{eq:cons-bnd}\\
  x_{k+1} &= d_k + A_k x_k + B_k u_k + \nu_k^{+} - \nu_k^{-} \label{eq:cons-dyn}\\
  \nabla_k p_k &\ge d_{\text{safe}} - \bar h_k + \nabla_k \bar p_k - \sigma_k
    \label{eq:cons-sdf}\\
  u_{\min} &\le u_k \le u_{\max}, \quad 
  r_{\text{disc}} \le x_k^{(1)} \le W - r_{\text{disc}} \label{eq:cons-inx}\\
  |x_k - \bar{x}_k| &\le \Delta x, \quad |u_k - \bar{u}_k| \le \Delta u
    \label{eq:cons-tr}
\end{align}
\end{subequations}
where $d_k = f_k - A_k \bar{x}_k - B_k \bar{u}_k$, with
$A_k = \partial f_{\text{edge}}/\partial x_k$,
$B_k = \partial f_{\text{edge}}/\partial u_k$ and
$f_k = f_{\text{edge}}(\bar x_k,\bar u_k)$ the true one-edge rollout and its
Jacobians. Here $d_{\text{safe}}$ is the required clearance, $r_{\text{disc}},W,H$ the robot
radius and workspace extents, and $\Delta x \in \mathbb{R}^5$, $\Delta u \in \mathbb{R}^2$ the per-dimension trust radii (the absolute values in \eqref{eq:cons-tr} are elementwise). With the risk constraints, we define a convex problem. The slacks $\nu^{\pm}_k$, $\sigma_T^{\pm}$ and $\sigma_k$ each carry an exact-$\ell_1$ penalty in the objective, making the subproblem feasible at all iterations. Problem \eqref{eq:objective} is solved iteratively using the commercial solver $\text{CLARABEL}$  \cite{goulart_clarabel_2024} until the convergence criteria are met. 

\section{Experiments and Results}
Trajectory-level risk is obtained by combining $\mathrm{RiskMap}$
values along the $K$ discretized trajectory nodes in
\eqref{eq:riskmap}, $R_{\text{traj}} = 1 - \prod_{k=1}^{K}\big(1 - \mathrm{RiskMap}(x_k)\big)$. To isolate the effect of risk-aware cost, we run identical experiments with and without the risk cost included in the AO-RRT objective.  Without it, the algorithm minimizes control cost and plans the trajectory directly between the two high-risk regions, incurring a joint collision risk of 0.94. Because SCP can only refine the trajectories in the same homotopy, it cannot completely escape the high-risk region. However, through this experiment, we demonstrate the effectiveness of SCP refinement. The joint collision risk after the AO-RRT solution decreases to 0.29 with a control cost of 18.6 units, shown in Fig. \ref{fig:pair-b}. Therefore, without the SCP, it was almost certain to collide with the obstacle, resulting in mission failure. With the risk cost active in AO-RRT shown in Fig. \ref{fig:pair-a}, the planned trajectory is in the neighborhood of low-uncertainty obstacles. Therefore, the SCP refinement optimizes for control cost and smooths the trajectory around low-uncertainty obstacles with a control cost of 18.4 and joint collision risk of $6\times10^{-3}$. 

The risk-aware cost in AO-RRT is a user-settable parameter; through this, the user can balance between an initial risk-aware solution being generated with higher control cost versus lower risk and vice versa. 

We validate the proposed approach through hardware experiments conducted on a Leo rover equipped with a ZED 2i stereo camera and a Jetson Orin AGX on granular terrain. Following the construction of a point cloud of the obstacle-rich environment, obstacles are identified, and uncertainties are allocated to construct the risk map. Following the construction of the risk map, the planner is used to obtain a dynamically feasible trajectory, which is then tracked by a contraction-based controller assuming unicycle dynamics using the pose estimated obtained from the Vicon motion capture system. We observe that despite terrain-induced wheel slip, the contraction controller \cite{126006} using slip-adjusted feedforward velocities tracks the trajectory with an RMS error of $6\text{cm}$.

\section{Conclusion}
In this work, we present a risk-aware motion planning method that generates dynamically feasible trajectories for a robotic agent in an uncertainty-rich environment and verify on hardware. Through numerical and hardware experiments, we show that SCP after risk-neutral AO-RRT reduces the joint collision risk by $\sim$69\%, and risk-aware AO-RRT generates an initial trajectory solution with $\sim$97\% less risk with comparable control cost. Our work provides a safer and risk-aware solution for the motion planning problem in planetary and lunar robotics applications, which is critical to mission success.

\bibliographystyle{IEEEtran}
\bibliography{main}

@INPROCEEDINGS{126006,
  author={Kanayama, Y. and Kimura, Y. and Miyazaki, F. and Noguchi, T.},
  booktitle={Proceedings., IEEE International Conference on Robotics and Automation}, 
  title={A stable tracking control method for an autonomous mobile robot}, 
  year={1990},
  volume={},
  number={},
  pages={384-389 vol.1},
  doi={10.1109/ROBOT.1990.126006}}

@article{hauser_asymptotically_2016,
	title = {Asymptotically {Optimal} {Planning} by {Feasible} {Kinodynamic} {Planning} in a {State}–{Cost} {Space}},
	volume = {32},
	copyright = {https://ieeexplore.ieee.org/Xplorehelp/downloads/license-information/IEEE.html},
	issn = {1552-3098, 1941-0468},
	doi = {10.1109/TRO.2016.2602363},
	number = {6},
	urldate = {2026-08-02},
	journal = {IEEE Transactions on Robotics},
	author = {Hauser, Kris and Zhou, Yilun},
	month = dec,
	year = {2016},
	pages = {1431--1443},
}

@article{lindemann_safe_2023,
	title = {Safe {Planning} in {Dynamic} {Environments} {Using} {Conformal} {Prediction}},
	volume = {8},
	copyright = {https://ieeexplore.ieee.org/Xplorehelp/downloads/license-information/IEEE.html},
	issn = {2377-3766, 2377-3774},
	doi = {10.1109/LRA.2023.3292071},
	number = {8},
	urldate = {2026-08-02},
	journal = {IEEE Robotics and Automation Letters},
	author = {Lindemann, Lars and Cleaveland, Matthew and Shim, Gihyun and Pappas, George J.},
	month = aug,
	year = {2023},
	pages = {5116--5123},
}

@article{singh_framework_2019,
	title = {A {Framework} for {Time}-{Consistent}, {Risk}-{Sensitive} {Model} {Predictive} {Control}: {Theory} and {Algorithms}},
	volume = {64},
	copyright = {https://ieeexplore.ieee.org/Xplorehelp/downloads/license-information/IEEE.html},
	issn = {0018-9286, 1558-2523, 2334-3303},
	shorttitle = {A {Framework} for {Time}-{Consistent}, {Risk}-{Sensitive} {Model} {Predictive} {Control}},
	doi = {10.1109/TAC.2018.2874704},
	number = {7},
	urldate = {2026-08-02},
	journal = {IEEE Transactions on Automatic Control},
	author = {Singh, Sumeet and Chow, Yinlam and Majumdar, Anirudha and Pavone, Marco},
	month = jul,
	year = {2019},
	pages = {2905--2912},
}

@misc{goulart_clarabel_2024,
	title = {Clarabel: {An} interior-point solver for conic programs with quadratic objectives},
	shorttitle = {Clarabel},
	doi = {10.48550/arXiv.2405.12762},
	urldate = {2026-08-02},
	publisher = {arXiv},
	author = {Goulart, Paul J. and Chen, Yuwen},
	month = may,
	year = {2024},
	note = {arXiv:2405.12762 [math.OC]},
}

@article{lupu_magicvfm_2025,
	title = {{MAGIC}$^{\textrm{{VFM}}}$ -{Meta}-{Learning} {Adaptation} for {Ground} {Interaction} {Control} {With} {Visual} {Foundation} {Models}},
	volume = {41},
	copyright = {https://ieeexplore.ieee.org/Xplorehelp/downloads/license-information/IEEE.html},
	issn = {1552-3098, 1941-0468},
	doi = {10.1109/TRO.2024.3475212},
	urldate = {2026-08-02},
	journal = {IEEE Transactions on Robotics},
	author = {Lupu, Elena Sorina and Xie, Fengze and Preiss, James Alan and Alindogan, Jedidiah and Anderson, Matthew and Chung, Soon-Jo},
	year = {2025},
	pages = {180--199},
}

@article{rockafellar2000optimization,
  title={Optimization of conditional value-at-risk},
  author={Rockafellar, R Tyrrell and Uryasev, Stanislav and others},
  journal={Journal of risk},
  volume={2},
  pages={21--42},
  year={2000}
}

@article{dixit2024step,
  title={Step: Stochastic traversability evaluation and planning for risk-aware navigation; results from the darpa subterranean challenge},
  author={Dixit, Anushri and Fan, David D and Otsu, Kyohei and Dey, Sharmita and Agha-Mohammadi, Ali-Akbar and Burdick, Joel},
  journal={Field Robotics},
  volume={4},
  pages={182--210},
  year={2024},
  publisher={FRPS}
}

@ARTICLE{10606099,
  author={Cai, Xiaoyi and Ancha, Siddharth and Sharma, Lakshay and Osteen, Philip R. and Bucher, Bernadette and Phillips, Stephen and Wang, Jiuguang and Everett, Michael and Roy, Nicholas and How, Jonathan P.},
  journal={IEEE Transactions on Robotics}, 
  title={EVORA: Deep Evidential Traversability Learning for Risk-Aware Off-Road Autonomy}, 
  year={2024},
  volume={40},
  number={},
  pages={3756-3777},
  doi={10.1109/TRO.2024.3431828}}

@ARTICLE{8767973,
  author={Hakobyan, Astghik and Kim, Gyeong Chan and Yang, Insoon},
  journal={IEEE Robotics and Automation Letters}, 
  title={Risk-Aware Motion Planning and Control Using CVaR-Constrained Optimization}, 
  year={2019},
  volume={4},
  number={4},
  pages={3924-3931},
  doi={10.1109/LRA.2019.2929980}}

@incollection{majumdar2020risk,
  title={How Should a Robot Assess Risk? Towards an Axiomatic Theory of Risk in Robotics},
  author={Majumdar, Anirudha and Pavone, Marco},
  booktitle={Robotics Research},
  editor={Amato, Nancy and Hager, Gregory and Thomas, Shelby and Torres-Torriti, Miguel},
  volume={10},
  series={Springer Proceedings in Advanced Robotics},
  year={2020},
  publisher={Springer, Cham},
  doi={10.1007/978-3-030-28619-4_10}
}
\end{document}